\documentclass[conference]{IEEEtran}
  \IEEEoverridecommandlockouts
  
  \usepackage{cite}
  \usepackage{amsmath,amssymb,amsfonts}
  \usepackage{algorithmic}
  \usepackage{graphicx}
  \usepackage{textcomp}
  \usepackage{subfigure}
  \usepackage{xcolor}
  \usepackage{bm}
  \usepackage{amsfonts}
  \usepackage{algorithm}
  \usepackage{algorithmic}
  \usepackage{multirow}
  \usepackage{verbatim}
  
  \def\BibTeX{{\rm B\kern-.05em{\sc i\kern-.025em b}\kern-.08em
      T\kern-.1667em\lower.7ex\hbox{E}\kern-.125emX}}
  \begin{document}
  
  \title{Approximate Random Dropout for DNN training acceleration in GPGPU\\
  }
  
  \author{
          Zhuoran~Song,
          Ru~Wang,
          Rudong~Yu,
          Hongru~Huang,
          Zhenghao~Peng,
          Li~Jiang*}

  \maketitle
  \begin{abstract}
  The training phases of Deep neural network~(DNN) consumes enormous processing time and energy. Compression techniques utilizing the sparsity of DNNs can effectively accelerate the inference phase of DNNs. However, it can be hardly used in the training phase because the training phase involves dense matrix-multiplication using General Purpose Computation on Graphics Processors (GPGPU), which endorse regular and structural data layout.
In this paper, we propose the Approximate Random Dropout that replaces the conventional random dropout of neurons and synapses with a regular and predefined patterns to eliminate the unnecessary computation and data access. To compensate the potential performance loss we develop a SGD-based Search Algorithm to produce the distribution of dropout patterns.
We prove our approach is statistically equivalent to the previous dropout method.
Experiments results on MLP and LSTM using well-known benchmarks show that the proposed Approximate Random Dropout can reduce the training time by $20\%$-$77\%$ ($19\%$-$60\%$) when dropout rate is $0.3$-$0.7$ on MLP (LSTM) with marginal accuracy drop.
  \end{abstract}
  

  \section{Introduction}\vspace{-5pt}
Deep Neural Networks~(DNNs) have emerged as critical technologies to solve various complicated problems~\cite{Luo2017Understanding, Young2017Recent}. The inference of DNNs is computational expensive and memory intensive and therefore has an urgent need for acceleration before we can fully embrace DNNs in the power-limited devices. Extensive works are proposed to reduce the computation by compressing the size of synaptic weights, such as weight pruning~\cite{Han2016Deep,Ding2017CirCNN}, quantization~\cite{Zhou2017Incremental,Gong2014Compressing,Leng2017Extremely}, low rank~\cite{Jaderberg2014Speeding,Ioannou2015Training} and Compact Network Design~\cite{Zhang2017ShuffleNet}. 
The above compression techniques may require retraining the DNN with limited accuracy loss~($<1\%$). 
The success of these techniques relies on the sparsity and plasticity of DNNs, however, cannot directly apply to the training phase of DNNs.

The training phase, involving the back-propagation through the network to update the weights, demands three-times more computation effort. GPGPU is suitable for such task which is attributed to the superior parallelism for large matrix multiplication~\cite{C2009Training, Puri2010Training}. Extensive works propose to accelerate the training phase on the distributed GPU-based system~\cite{Wen2017TernGrad,Zhang2016Staleness}.
Other works~\cite{Sun2017meProp, K2017Flexpoint} focus on accelerating the training phase using gradient pruning and weight quantization, respectively.

Random Dropout technique addresses the over-fitting problem and is widely used in MLP and LSTM training. The most common method~\cite{Srivastava2014Dropout} randomly dropping some neurons of each layer in every training iteration, while the other (DropConnect~\cite{Wan2013Regularization}) aims the same goal by randomly dropping some synapses connections between layers, namely some elements in weight matrix.
Theoretically, we can reduce the number of multiplication to $30\%$-$70\%$ if we can skip the calculation of all the dropped neurons or synapses while the dropout rate changes from 0.3 to 0.7.
However, such tremendous saving of multiplication as well as the corresponding data access is hard to exploit. Because the neurons or synapses are randomly and irregularly dropped following the Bernoulli distribution. Such irregularity prevents the GPGPU's single instruction multiple threads~(SIMT) architecture to skip the unnecessary multiplication and memory access.

Therefore, in this work, we replace the random dropout with two types of regular dropout patterns to make the choices of dropped neurons or synapses predictable, which allow GPGPU to skip calculation of those dropped neurons or synapses. We further developed an SGD-based Search Algorithm to produce the distribution $\mathcal K$ of dropout patterns such that the dropout rate of each neuron is approximately subjected to a Bernoulli distribution (We provide a brief proof). In each iteration, we sample a dropout pattern subjected to $\mathcal K$ and then eliminate the redundant computation by omitting the dropped data during the hardware allocation.
Our experiments show that applying Approximate Random Dropout during training can reduce the training time by $20\%$-$77\%$ ($19\%$-$60\%$) when dropout rate is $0.3$-$0.7$ on MLP (LSTM) with less than $0.5\%$ accuracy loss. We find that when the batch size increases, the speedup rate increases with accuracy of neural network declines.

The reminder of the paper is organized as follows: Section~\ref{sect:related} introduces the related works and motivates this paper. Section~\ref{sect:dropout} describes the proposed Approximate Random Dropout Technique. Experiments are shown in Section~\ref{sect:experiments}. Section~\ref{sect:conclusion} concludes this paper.
  \section{Background}\label{sect:related}
\subsection{Accelerating DNN inference and training}
There are considerable works pitch into accelerating inference of DNN by leveraging the sparsity of DNN~\cite{Zhang2017ShuffleNet,Wen2017Learning,Zhou2017Incremental,Ding2017CirCNN}.
Han et al.~\cite{Han2016Deep} prune synaptic weights which are close to zero and then retrain the DNN to maintain the classification accuracy. 
The zero weights are then encoded and moved onto the on-chip memory.
Special decoder is deployed in the accelerator to decode the zero weights and skip the computation.
Consequently, above methods can only benefit ASIC/FPGA based DNN accelerator instead of GPU.
Jaderberg et al.~\cite{Jaderberg2014Speeding} and Ioannou et al.~\cite{Ioannou2015Training} use low-rank representations to create computationally efficient neural networks. These methods cannot be used in training phase because of the subtle change of the weights degrades the convergence and accuracy of the training phase.


Extensive works propose to accelerate the training phase on the distributed GPU-based system~\cite{Wen2017TernGrad,Zhang2016Staleness}. Wen et al.~\cite{Wen2017TernGrad} propose to use ternary gradients to accelerate distributed deep learning in data parallelism. Zhang et al.~\cite{Zhang2016Staleness} propose a variant of the asynchronous SGD algorithm to guarantee the convergence of this algorithm and accelerate the training in a distributed system. 
Other works are relative to the acceleration in the training process using gradient pruning and weight quantization.
Kster et al.~\cite{K2017Flexpoint} share the exponent part of the binary coding of the weights and thereby convert floating-point operations to fixed-point integer operations. Noted that this work is compatible with ours and we leave this topic to further research.
Sun et al.~\cite{Sun2017meProp} prune those comparatively small gradients to speed up training phase. However, their work focuses on software-level optimization and thus yields marginal training acceleration while this work enable computation reduction on hardware-level.

\subsection{Basics of the GPGPU}\label{ssect:gpu}
GPGPU is commonly used for DNN training. 
It is composed of dozens of streaming multiprocessors~(SMs).
Each SM consists of single instruction multiple threads~(SIMT) cores and a group of on-chip memories including register file, shared memory, L1D cache and etc.
Each SM manages and executes multi-threads on it. Those threads are clustered into warps~(e.g., 32 threads in NVIDIA GPU), executing the same instruction at the same time. 
Thus, the branch divergence occurs when programmers write conditional branch (if-else).

Shared memory is a performance-critical on-chip memory. 
The latency of accessing the global memory (DRAM) is roughly 100x higher than that of accessing the shared memory. Hence, reducing the frequency of accessing global memory is critical for performance.
The capacity of the shared memory per block is 48KB in Nvidia GTX 1080Ti, which is much smaller than the capacity of the global memory. Therefore, reducing the superfluous data in shared memory is also important.

The key purpose of this work is to reduce the scale of matrices, by which we can reduce the access frequency of the shared memory and the global memory as well as the computation effort to accelerate the training.

\subsection{Random Dropout}\label{ssect:dropout}
Random dropout is widely used to prevent over-fitting. It randomly omits part of the neurons~\cite{Srivastava2014Dropout} or synapses~\cite{Wan2013Regularization} on each training iteration.
The probability of a neuron or a synapses to be dropped is subjected to a Bernoulli distribution parameterized with \emph{dropout rate}~\cite{Wan2013Regularization,Wen2017Learning}. In a nutshell, the main reason why random dropout can effectively prevent over-fitting is that it generates adequate different sub-models to learn diverse features during the training process and ensembles those sub-models to maximize the capability of DNN for inference.

\begin{figure}[htbp]
    \centering
    \begin{minipage}{0.45\linewidth}
    \centering
    \includegraphics[width=4cm,height=3.5cm]{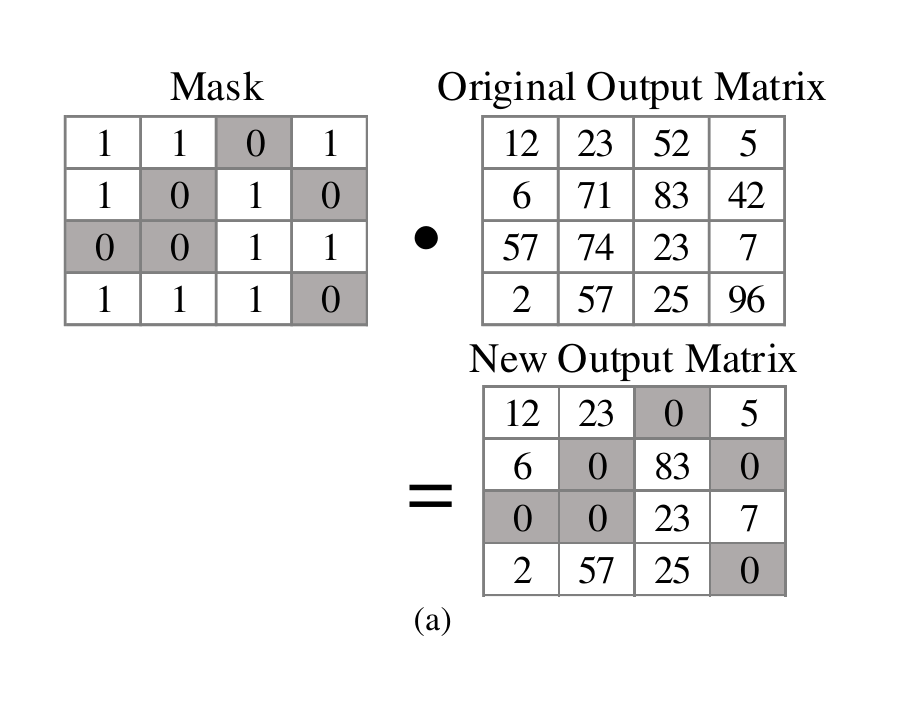}
    \end{minipage}
    \begin{minipage}{0.45\linewidth}
    \centering
    \includegraphics[width=4cm,height=3.5cm]{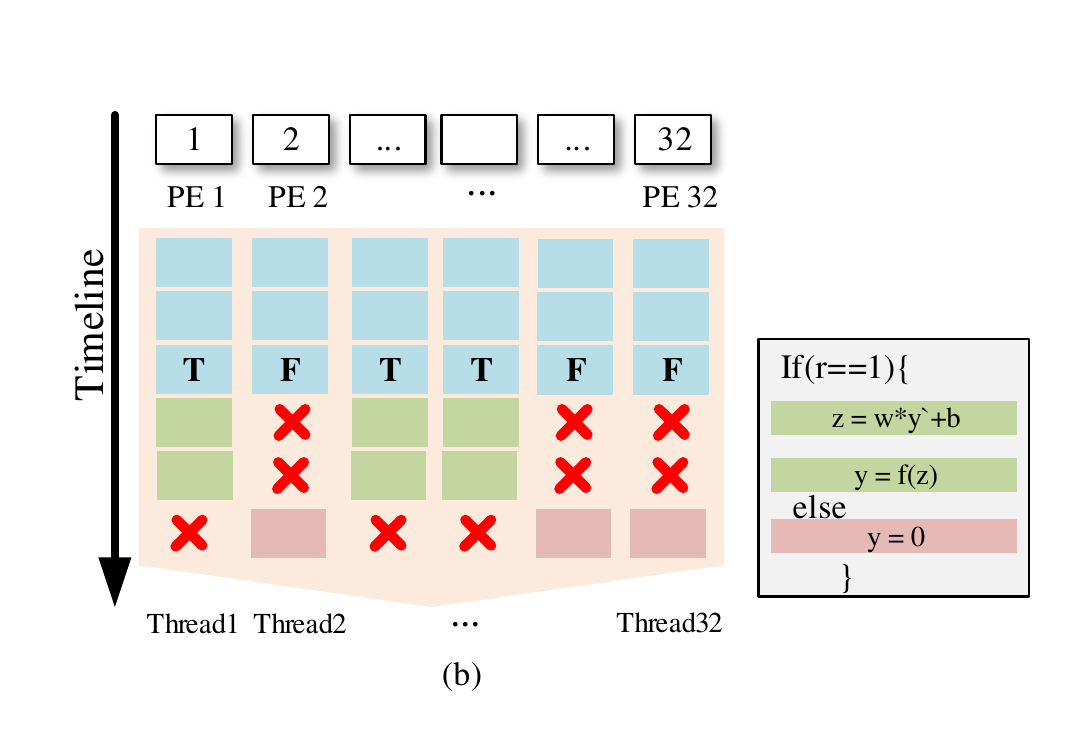}
    \end{minipage}
    \caption{(a) Implementation of Random Dropout in Forward propagation; (b) GPU divergence happened when directly skipping redundant calculation.}
    \label{FCTD}
\end{figure}

Existing machine learning frameworks, like Caffe~\cite{Jia2014Caffe} and Tensorflow~\cite{Abadi2016TensorFlow}, all adopt the dropout technique. For each layer in the forward propagation, the output matrix is computed and thereafter element-wisely multiplied by a mask matrix composed of randomly generated $0$s and $1$s, as shown in Fig.~\ref{FCTD}(a). Similarly, in back-propagation, they first calculate the derivatives of the output matrix. The resulting derivative matrix then multiplies by the same mask matrix.

A question arises: why not skipping the calculation of those dropped neurons to reduce the redundant time spent on the matrix multiplication and the data movement.
Intuitively, we can write conditional branch~(if - else) to skip the redundant calculation. However, such conditional branches incur branch divergence in GPU, which is a great hurdle for performance.
As shown in Fig~\ref{FCTD}(b), `T' denotes the threads that are satisfying the conditions ($r = 1$) and executing the green function($z = w*y'+b$ and $y = f(z)$), while `F' refers to those executing the red function($y = 0$). 
In GPGPU’s SIMT architecture, the red threads have to wait for the green threads. Thus, some process elements~(PEs) are idle, represented by the red cross. The total execution time is not reduced~(even increased) due to the branch divergence. Thus, it is non-trivial to exploit the dropout for speedup in GPU.
  \section{Approximate Random Dropout}\label{sect:dropout}
\begin{figure}[tb]
\centering\vspace{-10pt}
\includegraphics[width=0.7\linewidth]{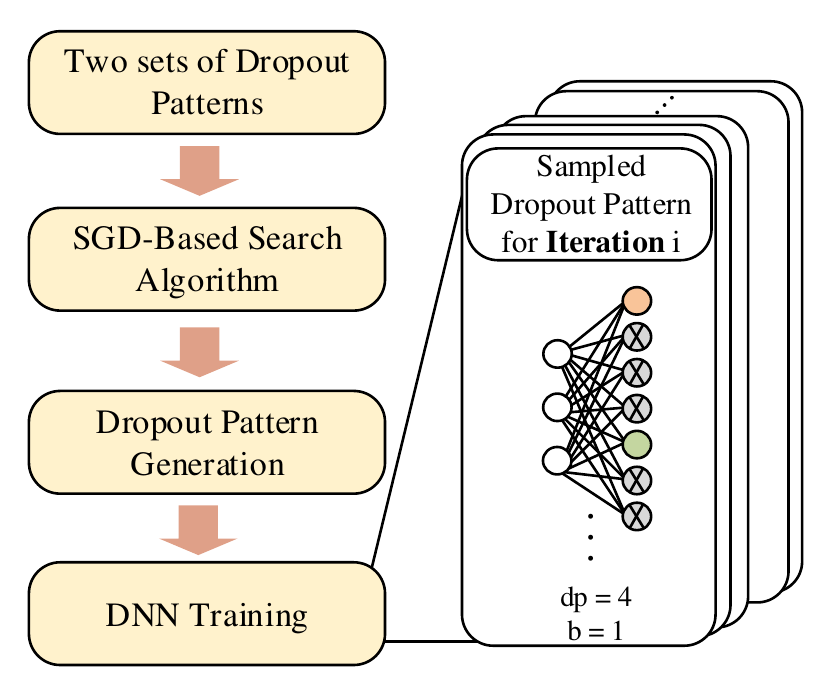}\vspace{-10pt}
\caption{Overview of the Approximate Random Drouput.}
\label{overview}\vspace{-10pt}
\end{figure}

\begin{figure*}[!tb]
\centering
\subfigure[Row-based Dropout Pattern]{
\includegraphics[width=0.45\linewidth]{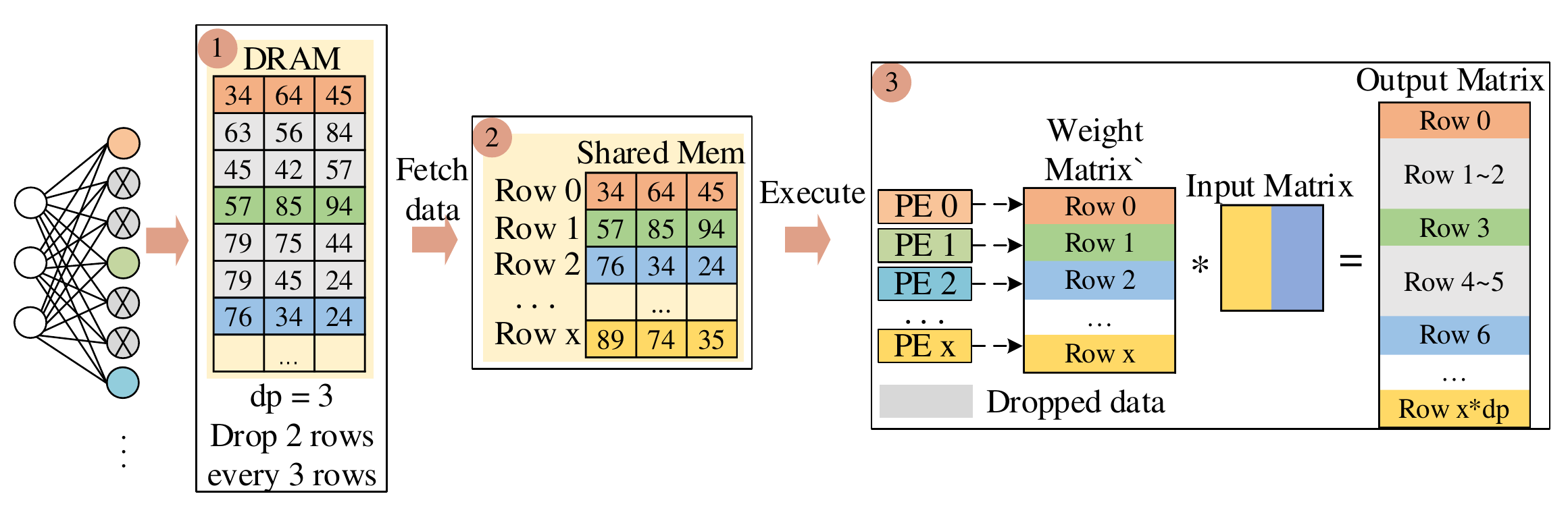}\vspace{-10pt}
\label{pattern}\vspace{-10pt}
}
\subfigure[Tile-based Dropout Pattern]{
\includegraphics[width=0.45\linewidth]{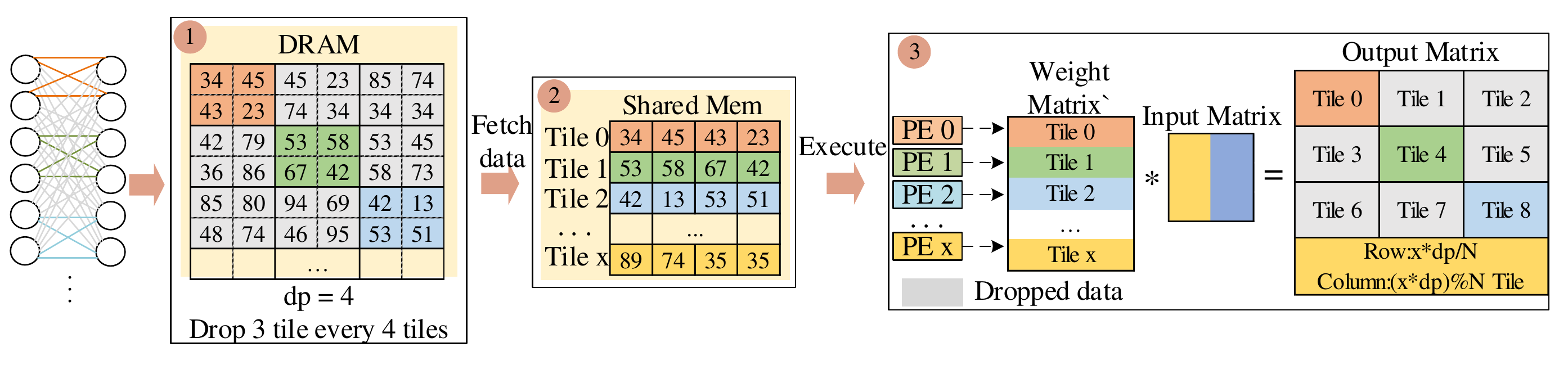}\vspace{-10pt}
\label{tile_pattern}\vspace{-10pt}
}
\caption{GPU hardware allocation during training.}
\end{figure*}
The key idea of accelerating the DNN training is to reduce the scale of matrices involved in multiplication and avoid the brunch divergence of GPU. However, the randomness in conventional dropout methods hamper the scale reduction.

In this work, we define \emph{dropout pattern} as the combination of dropped neurons in each training iteration. As shown in Fig.~\ref{overview}, we design two sets of regular dropout pattern and replace the random dropout with a sampled dropout pattern sampled from them. Resulted from the replacement, we can forecast which neurons or synapses to be dropped and thereafter assist GPU to skip the calculation and data access of the dropped neurons without incurring the divergence. We modify the caffe source code to reduce the scale of matrices which become feasible due to the predictable dropout.

However, the loss of randomness induced by the regular dropout patterns increases the risk of over-fitting the DNN.
To cope with this issue, we further develop a Stochastic Gradient Decedent~(SGD) based Search Algorithm~(see section~\ref{ssect:sgd}), to find a distribution of all possible dropout pattern such that the probability distributions of each neuron or synapse being dropped between our method and conventional method is equivalent. We provide a brief proof of that.


In this section, based on the computation characteristic of GPU, we firstly propose two sets of Dropout Patterns---Row-based Dropout Pattern (RDP) and Tile-based Dropout Pattern (TDP)---and then analyze the mechanisms of the reduction of the time of computation and data access. After that, we introduce our SGD-based Search Algorithm which produce a distribution of possible dropout patterns as well as the dropout pattern generation procedure in each iteration.

\subsection{Row-based Dropout Pattern}

In conventional dropout method, the computation relative to a dropped neuron is the multiplication between zero and the correspondent row in the weight matrix of next layer to shrink the scale of matrix, in RDP, we drop the whole row in the weight matrix, which is equivalent to drop all the synapses of a specified neuron.

Concretely, RDP is parameterized by two scalar $dp$ and bias $b$ as follow: we uniformly choose a bias $b\in \{1,..., dp\}$ and drop all rows in the weight matrix whose indices satisfy 
\begin{equation}
    i: (i-b) \bmod dp=0
\end{equation}
Consequently, $(i-1)/(i)$ of the neurons are dropped. For instance (the left of Fig.~\ref{pattern}), when $dp$=$3$, $b$=$1$, starting from the first row, we drop two rows~(i.e., neurons) in every successive three rows~(neurons) in the weight matrix.

Given the size of the output matrix as $M \times N$, the maximum $dp$ is $dp_{max} = M$, and the maximum number of the sub-models is $\sum_{i=1}^{M} i=(M+1)/2$ considering the number of possible bias is $i$ when $dp=i$.

The execution processes in GPU is also shown in Fig.~\ref{pattern}. DRAM stores the whole weight matrix~(as shown in step 1); the gray block denotes the rows of weight matrix correspondent to the dropped neurons. We write the kernel function to prevent GPU from fetching those dropped data into shared memory~(as shown in step 2) and build two compact matrices (input matrix and weight matrix) for next step. After data fetch, every PE multiplies one row of the weight matrix by the whole input matrix. Thus, only $\frac{1}{dp}$ of the original weight matrix as well as the input matrix is fetched and calculated. The resulting rows fill $1\times dp$ rows in the Output Matrix using the same pattern. The rest $\frac{dp-1}{dp}$ of the Output Matrix is set to zero by default. Note that the RDP is agnostic to the matrix-multiplication algorithm as it temporarily compresses the matrices into a compact layout. Therefore, RDP can comply to any optimization method for matrix multiplication.

\subsection{Tile-based Dropout Pattern (TDP)}

Tile is a sub-matrix in weight matrix and contains multiple synapses connections. We use tiles as the unit to drop rather than synapse~\cite{Wan2013Regularization} (namely the size of tiles is $1$) for the purpose of regularity.
TDP is also parameterized by $dp$ and bias $b$. $dp-1$ tiles are dropped in every $dp$ tiles, resulting in $\frac{dp-1}{dp}$ of synapses connections being dropped.
As shown in the left of Fig.~\ref{tile_pattern}, when $dp=4, b=1$, starting from first tile, we drop 3 tiles in every 4 successive tiles.

TDP have similar procedure compare to RDP but different in: (1)TDP fetches non-dropped tiles into the shared memory rather than rows, and builds two compact matrices. (2) each PE conduct the multiplication of one tile of compact weight matrix and the corresponding tile of compact input matrix, according to their PE index. In the right of Fig.~\ref{tile_pattern}, GPU only conduct multiplication of two compact matrices whose scale is $\frac{1}{4}$ of the original scale.
This Dropout Pattern can naturally work with Tiling Method~\cite{ryoo2008optimization} in matrix multiplication, which is an essential optimization technique.

Given the size of the output matrix $M \times N$, the size of the tile $x \times y$, the maximum $dp$ is $dp_{max} = \left \lfloor {M}/{x} \right \rfloor \times \left \lfloor {N}/{y} \right \rfloor$ and the maximum number of sub-models is $(1+dp_{max})/2$. 
TDP can generate more sub-models than RDP, when $N$ is roughly greater than $x\times y$.

The choice of tile size is critical: the smaller size of the tile, the more number of Dropout Patterns as well as sub-models, while small tile leads to fine-grained control.
Under such circumstances, the size of tile is set to be $32\times 32$ to balance the maximization of the number of sub-models and avoiding shared memory's bank conflict since the shared memory has 32 banks in NVIDIA GPU.

A typical training process is composed of three steps: fully connected layer computation, activation layer computation and dropout layer computation using the mask matrix. 
After applying the Dropout Pattern with $dp = 2$, we only need to spend half of the time for fully connected layer computing and skip the dropout layer computing. 
Consequently, given the dropout pattern, the time spending on training can be overtly reduced.

\subsection{SGD-based Search Algorithm for Dropout Pattern Distribution}\label{ssect:sgd}

For each iteration in training procedure, only one regular dropout pattern is applied to the network. In order to approximate the traditional dropout process~\cite{Srivastava2014Dropout}, the dropout pattern we choose in each iteration should satisfy that: (1)the dropout probability of each neuron should subject to a given Bernoulli distribution, and (2)different sub-models derived from that series of dropout patterns should be adequate.

Therefore, we propose an efficient SGD-based Search Algorithm to find a dropout pattern distribution from which the dropout pattern sampled satisfy the demands. SGD consumes tractable time and is convenient in optimizing the continuous variables.
More specifically, the algorithm obtains a probability distribution $\mathcal K=\{k_i\}_{i=1}^{dp_{max}}$ which contains the probability $k_i$ of each possible Dropout Pattern $i\in \{1,2,..., dp_{max}\}$, who is subjected to $\sum_{i=1}^{dp_{max}}{k_i} = 1$. 

Here we define the \emph{global dropout rate} as the proportion of neurons or synapses who are set to zero. Noted that the global dropout rate is different from the conventional dropout rate which refer to the probability of a single neuron or synapse to be dropped. However, we prove that within our approach the two dropout rate are statistically equivalent.

\begin{algorithm}[tb]
\begin{algorithmic}[1]
\caption{SGD-based Search Algorithm}
\label{algo}
\REQUIRE ~~\\
the target global dropout rate $p$, the maximum number of dropout pattern $N$, hyper parameters $\lambda_1$, $\lambda_2$ ($\lambda_1 + \lambda_2 = 1$).
\ENSURE ~~\\
A distribution $\mathcal K$ for different dropout patterns.
\STATE Initialize a $N$ dimension row vector $v$;
\STATE Initialize a $N$ dimension constant row vector $p_u$ as [0, 1/2, 2/3, ..., (N-1)/N];
\WHILE{$\left| \Delta loss\right| \geq threshold$}
\STATE $d = softmax(v)$;
\STATE $E_p = || d^T\cdot p_u  - p||^2_2$;
\STATE $E_n = \frac{1}{N}\sum_{i=1}^{N} d_i \log d_i$;
\STATE $Loss = \lambda_1E_p + \lambda_2E_n$;
\STATE Calculate gradients of loss over $v$;
\STATE Update $v$ with gradient descent;
\ENDWHILE
\STATE Return $d$;
\end{algorithmic}
\end{algorithm}

Given the target global dropout rate $p$, and the maximum $dp$ as $N$, we use Algorithm~\ref{algo} to search for desired distribution $\mathcal K$.
A vector $v$ with length $N$ is first arbitrary initialized (line 1) and the $softmax(v)$ serve as the final probability distribution of each dropout pattern (line 4). Then we setup a constant vector $p_u = \{\frac{i-1}{i}\}_{i=1}^{N}$ whose element denotes the global dropout rate of a given dropout pattern. Therefore, $d^T\cdot p_u$ is the expected global dropout rate and the difference between it and the target global dropout rate is our optimization objective (line 5). To enforce $d$ to be dense and to produce more diversified sub-models, the negative information entropy of $d$ is added to the loss (line 6, 7).
Then the algorithm uses SGD algorithm to update $v$ (line 8, 9) and return the distribution $d\in [0,1]^{N}$ when the loss is stuck.
By the loss function we defined in line 7, the algorithm aims at finding a distribution $\mathcal K$ that (1)make the global dropout rate equal to required value $p$ and (2)maximize the sub-models diversity.

\subsection{Dropout Pattern Generation}

The acquired distribution $\mathcal K$ is then used to sample dropout pattern in each iteration. Concretely, in each iteration, we randomly sample a dropout pattern (parameterized by $dp$ and $b$) subjected to the distribution $\mathcal K$, and then uniformly choose a bias $b\in \{1,..., dp\}$. Dropout pattern is then determined.

In our method, global dropout rate is statistically equivalent to the single neurons or synapse dropout rate. For each neuron or synapse, the probability of it to be dropped (conventional dropout rate) is:
\begin{equation}
p_{n}=\sum_{i=1}^{dp_{max}}p_b k_i=\sum_{i=1}^{dp_{max}}\frac{i-1}{i}k_i
\end{equation}
The global dropout rate of $\mathcal K$ is:
\begin{equation}
p_g = d^T\cdot p_u = \sum_{i=1}^{dp_{max}} k_i\frac{i-1}{i} \approx p
\end{equation}
Therefore, in terms of the whole training process, the dropout rate $p_n$ of a single neurons or synapse is equal to the global dropout rate $p_g$ and thus is approximately equal to the target dropout rate $p$ by the SGD-based Search Algorithm.
  \section{Experiments}\label{sect:experiments}
To evaluate the effectiveness of proposed approximate random dropout, we compare it with the conventional dropout technique in terms of the DNN accuracy and the training time.
In section~\ref{ssect:dif_drop}, in order to explore the influence of the dropout rate on the performance of a specific 4-layer Multilayer perceptron~(MLP), we vary different dropout rate on a MLP. Note that the dropout rate in our method refer to the target dropout rate $p$ as described in Section~\ref{ssect:sgd}.
In section~\ref{ssect:dif_network}, we compare different MLPs with a specific dropout rate.
The data set we use with MLP is MNIST~\cite{Krizhevsky2009Learning}.
Long short-term memory(LSTM) neural network~\cite{Graves2012Long} is used in section~\ref{ssect:scaling_lstm} to verify the scalability of our method. The dataset we used with LSTM include a dictionary whose size is 8800, and the Penn Treebank~(PTB)~\cite{marcus1993building} data set which has long been a central data set for language modeling.
The experiment codes is implemented in Caffe~\cite{Jia2014Caffe} and use a single GTX1080Ti GPU to run.

\subsection{Comparison of different dropout rate}\label{ssect:dif_drop}
The structure of a specific 4-layer MLP is described as follow: the input layer is shaped according to the batch size; the output layer has 10 neurons for digit 0 to 9; the two hidden layers have 2048 neurons both. During training, we set the following hyper-parameter: the batch size is 128, the learning rate is 0.01, and momentum is 0.9.

\begin{figure}[htbp]
    \centering
    \begin{minipage}{0.45\linewidth}
    \centering
    \includegraphics[width=4cm,height=4cm]{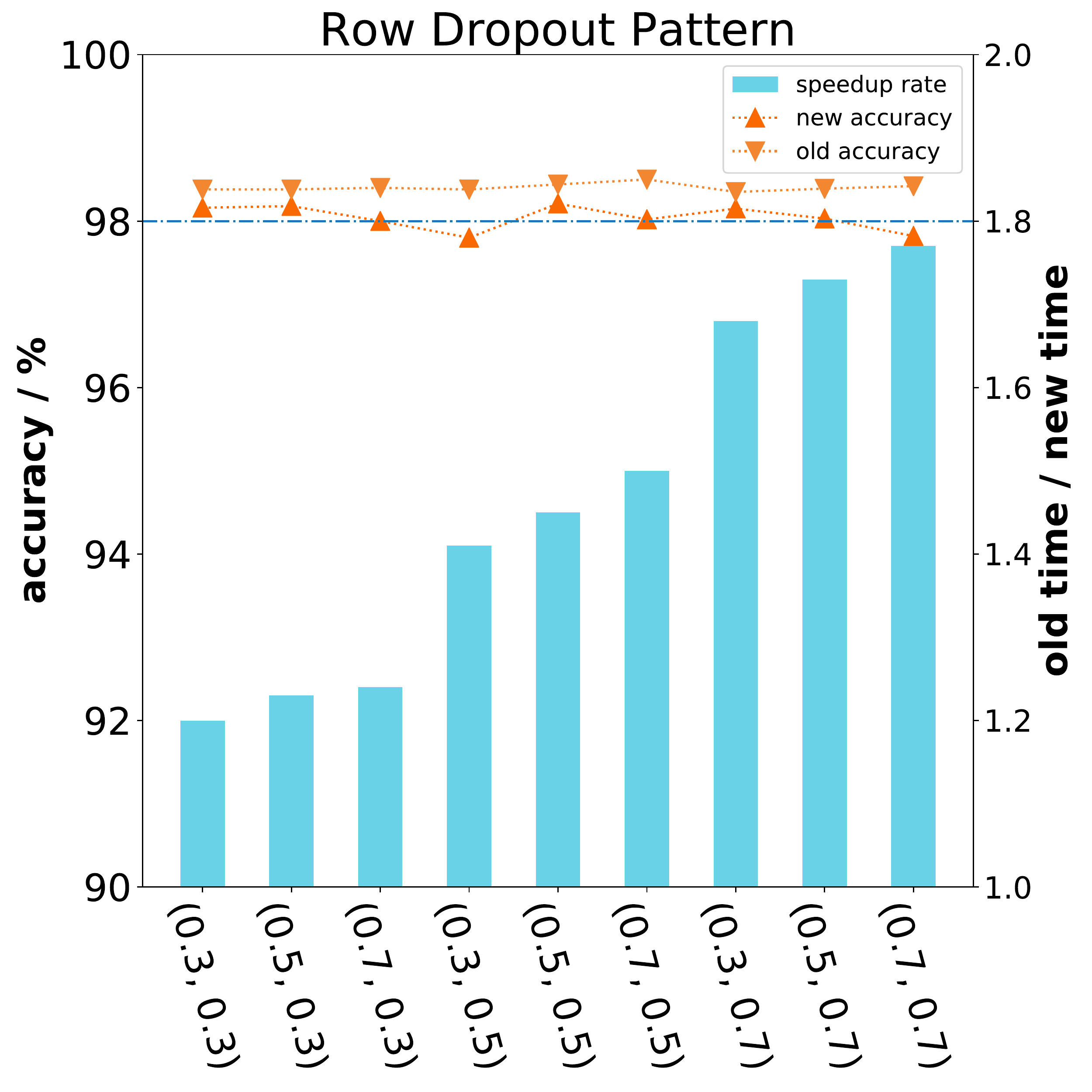}
    \end{minipage}
    \begin{minipage}{0.45\linewidth}
    \centering
    \includegraphics[width=4cm,height=4cm]{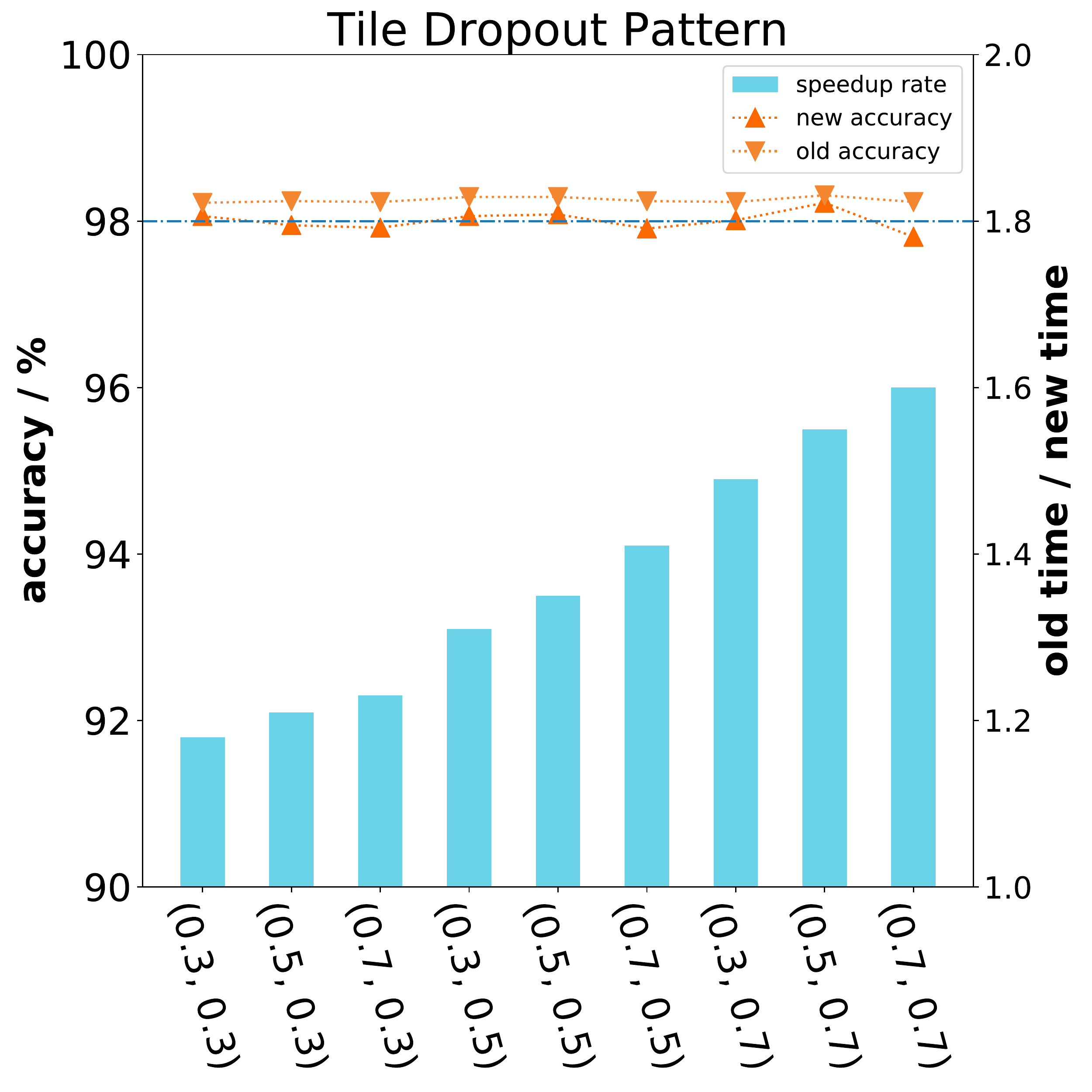}
    \end{minipage}
    \caption{Comparing different dropout rate combinations on specific network}
    \label{dif_drop}
\end{figure}

We vary the dropout rate from $(0.3, 0.3)$ to $(0.7,0.7)$ (two hidden layers may have varied dropout rate), and record the accuracy and training time for each dropout rate.
The comparison of two metrics of RDP and TDP against the conventional dropout are shown in Fig.~\ref{dif_drop}. The training time of conventional dropout is divided by the new training time of proposed approximate random dropout to obtain the speedup rate.

The results show that RDP can obtain $20\%-80\%$ speedup compared with the traditional dropout technique when the dropout rate varies between 0.3 and 0.7, which comply to our intuition as the amount of data that require no calculation expands with the increment of the dropout rate.
The speedup rate brought by TDP ranges from 1.18 to 1.6. The little slowdown is induced by the calculation of the nonzero positions in the output matrix before matrix multiplication.
The accuracy loss of these two classes of dropout patterns is less than $0.47\%$, which is the evadible concession to the speedup. TDP has less accuracy loss than RDP which can be attributed to the abundance of sub-models in TDP.

\subsection{Comparison of different networks}\label{ssect:dif_network}
We investigate the speedup in different MLP structures using a fixed dropout rate (0.7, 0.7).
Those MLPs have the same input and output layer as described in section~\ref{ssect:dif_drop}. Their hidden layer size is shown in Table~\ref{table_dif_network}. For instance, $1024\times64$ in the second column means the first and the second hidden layer's size are 1024 and 64, respectively.
The hyper-parameters of optimization algorithm follow above experiments.

From Table~\ref{table_dif_network}, the accuracy degradation is less than $0.5\%$. In some cases, the accuracy even increases.
Moreover, the speedup rate increases as the network size increases. Especially, in the case of $4096\times 4096$ network, both of the proposed dropout patterns reach a $2\times$ speedup.

\begin{table}[htbp]
\centering
\caption{Comparing different network with specific dropout rate}
\label{table_dif_network}
\begin{tabular}{|c|c|c|c|c|}
\hline
\multicolumn{1}{|l|}{\begin{tabular}[c]{@{}l@{}}Dropout \\ rate\end{tabular}} & \multicolumn{1}{l|}{\begin{tabular}[c]{@{}l@{}}Network \\ size\end{tabular}} & \multicolumn{1}{l|}{\begin{tabular}[c]{@{}l@{}}Dropout \\ pattern\end{tabular}} & \multicolumn{1}{l|}{Accuracy(its loss)} & \multicolumn{1}{l|}{\begin{tabular}[c]{@{}l@{}}Speedup \\ rate\end{tabular}} \\ \hline
\multirow{8}{*}{0.7}                                                           & \multirow{2}{*}{1024*64}                                                     & ROW                                                                             & 98.07\%(-0.42\%)                          & 1.27                                                                         \\ \cline{3-5}
                                                                               &                                                                              & TILE                                                                            & 98.11\%(-0.38\%)                          & 1.19                                                                         \\ \cline{2-5}
                                                                               & \multirow{2}{*}{1024*1024}                                                   & ROW                                                                             & 98.01\%(-0.35\%)                          & 1.45                                                                         \\ \cline{3-5}
                                                                               &                                                                              & TILE                                                                            & 98.15\%(-0.21\%)                          & 1.41                                                                         \\ \cline{2-5}
                                                                               & \multirow{2}{*}{2048*2048}                                                   & ROW                                                                             & 98.44\%(0.37\%)                           & 1.77                                                                         \\ \cline{3-5}
                                                                               &                                                                              & TILE                                                                            & 98.5\%(-0.31\%)                           & 1.60                                                                         \\ \cline{2-5}
                                                                               & \multirow{2}{*}{4096*4096}                                                   & ROW                                                                             & 98.00\%(-0.47\%)                          & 2.16                                                                         \\ \cline{3-5}
                                                                               &                                                                              & TILE                                                                            & 98.16\%(-0.31\%)                          & 1.95                                                                         \\ \hline
\end{tabular}
\end{table}

\subsection{Scaling to Long Short-Term Memory Model}\label{ssect:scaling_lstm}
We evaluate the speedup rate and the model performance on LSTM, which predicts the following word based on the given words. Each of the two hidden layers of LSTM contain 1500 neurons.
During training, we set the following hyper-parameters: the base learning rate is 1~(the base learning rate will gradually decrease), batch size is 20, the maximum epoch is 50, and the length of the sequence is 35.
It should be noted that the execution of LSTM is also performed as matrix multiplication, thus our proposed approximate dropout can be easily applied to LSTM.

As shown in Table~\ref{table_large_network}, the accuracy degradation is less than $1\%$. When dropout rate is increasing, the speedup rate increases without undermining the accuracy loss.

\begin{table}[htbp]
\centering
\caption{A dictionary data set which contains 8800 words on LSTM.}
\label{table_large_network}
\begin{tabular}{|l|l|l|l|l|}
\hline
\multicolumn{2}{|l|}{dropout rate}   & (0.3,0.3) & (0.5,0.5) & (0.7,0.7) \\ \hline
\multirow{3}{*}{accuracy} & original & 47.9\%    & 47.3\%    & 45.9\%    \\ \cline{2-5}
                          & ROW      & 46.9\%    & 46.0\%    & 44.5\%    \\ \cline{2-5}
                          & TILE     & 47.2\%    & 46.5\%    & 44.4\%    \\ \hline
\multirow{3}{*}{speedup}  & original & 1.0       & 1.0       & 1.0       \\ \cline{2-5}
                          & ROW      & 1.18      & 1.47      & 1.53      \\ \cline{2-5}
                          & TILE     & 1.18      & 1.43      & 1.49      \\ \hline
\end{tabular}
\end{table}

\begin{figure}[htbp]
\centering\vspace{-10pt}
\includegraphics[width=6cm]{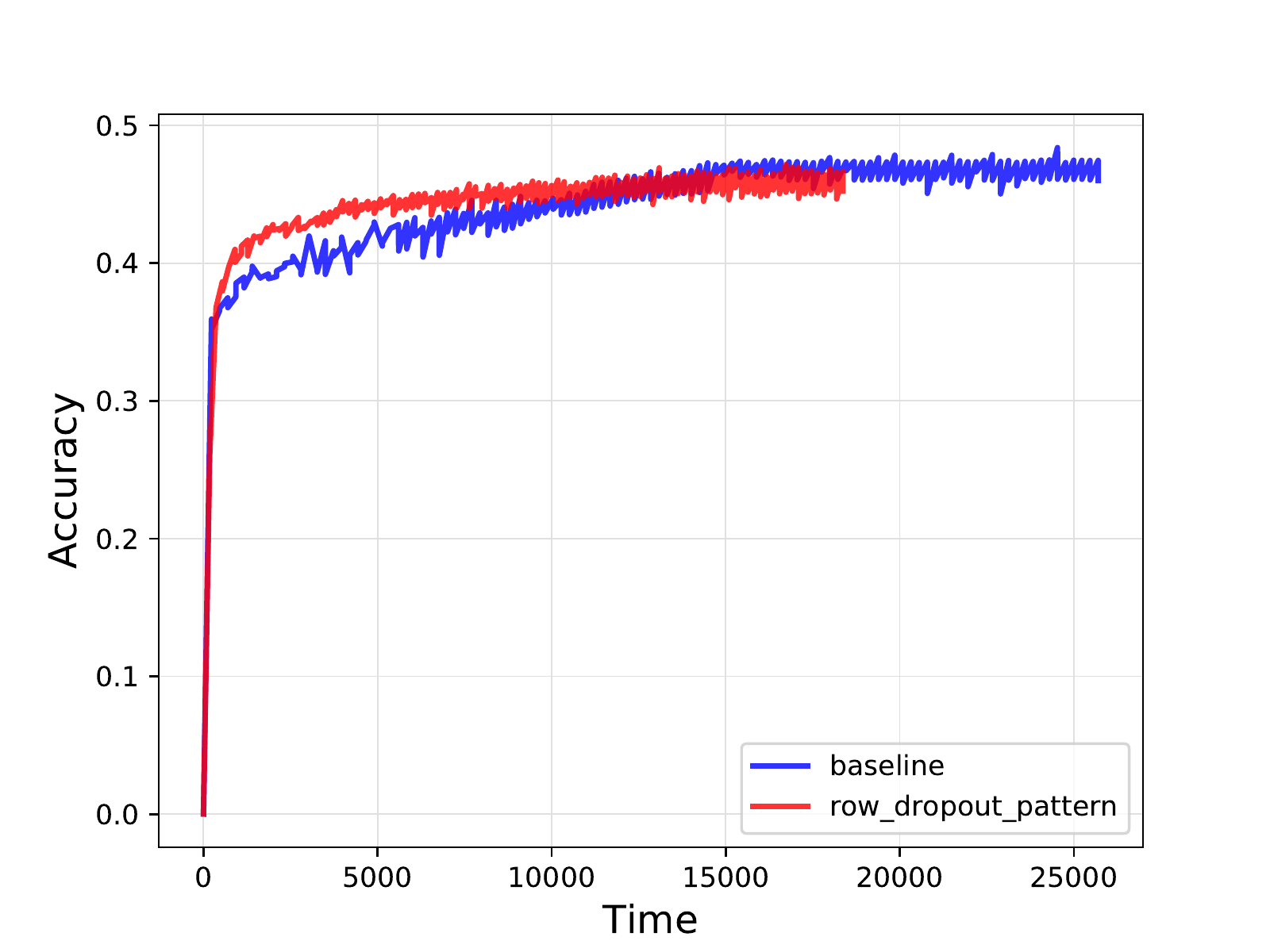}\vspace{-10pt}
\caption{The training process of RDP and traditional dropout.}
\label{train_log}\vspace{-10pt}
\end{figure}

To illustrate the effectiveness of the proposed method, we fix the dropout rate to $0.5$ and trace the accuracy of RDP until it's convergence. As shown in Fig.~\ref{train_log}, the red curve records our approximate random dropout training process; the blue one records the traditional dropout. The convergence of our method is eariler than the traditional dropout. Moreover, the smoothness of red curve indicates the approximate random dropout is helpful for the training process.

\begin{figure}[htbp]
    \centering
    \begin{minipage}{0.45\linewidth}
    \centering
    \includegraphics[width=4cm,height=3.5cm]{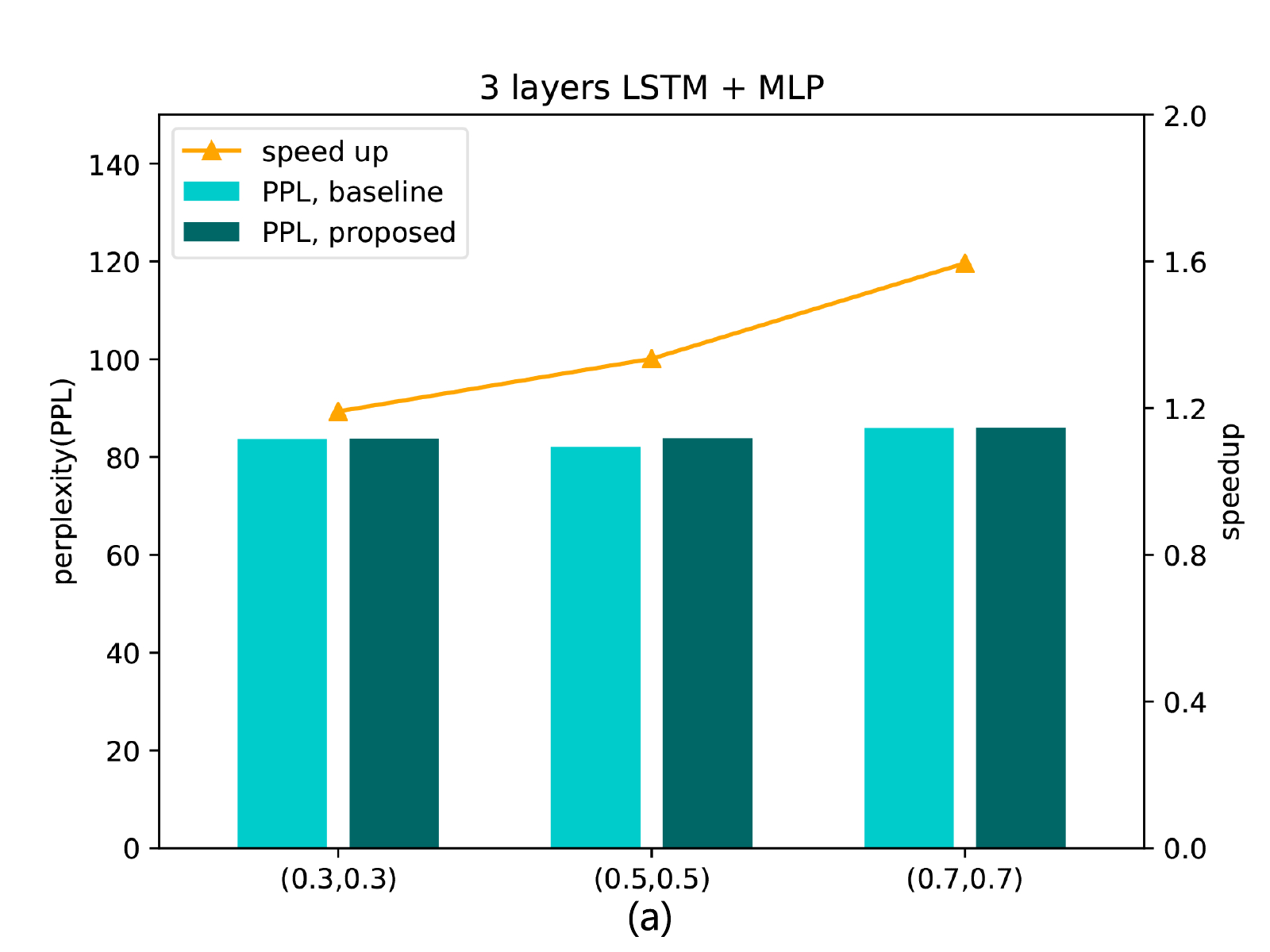}
    \end{minipage}
    \begin{minipage}{0.45\linewidth}
    \centering
    \includegraphics[width=4cm,height=3.5cm]{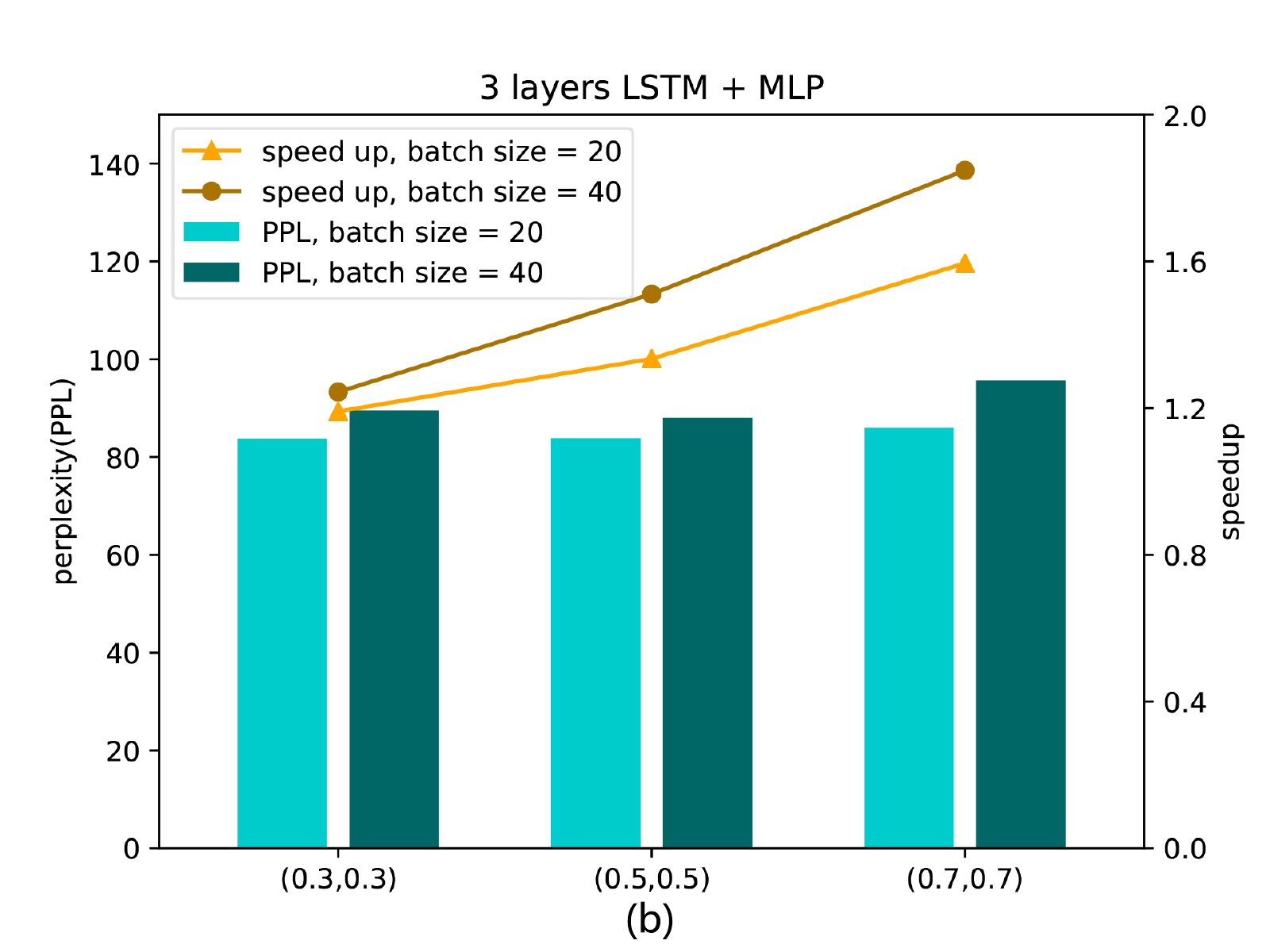}
    \end{minipage}
    \caption{Speedup rate and accuracy of Row approximate dropout on 3-layer LSTM.}
    \label{row_lstm}
\end{figure}

The result using the Penn Treebank data set~(PTB)~\cite{marcus1993building} on the 3-layer LSTM is shown in Fig.~\ref{row_lstm}(a). The test perplexity using RDP only increases $0.04$ given the dropout rate is 0.7, which further shows that our proposed approximate dropout algorithm can generate adequate sub-models for PTB data set.
Besides, when dropout rate increases from 0.3 to 0.7, the speedup rate also increases from $24\%$ to $85\%$.

We vary the batch size from 20 to 40.
Noted that SGD based search and data initialization are an one-time effort. When the batch size is increased, only the matrix operation and data transmission time increase accordingly. As shown in Fig.~\ref{row_lstm}(b), the speedup rate increases when batch size increases.
However, since one dropout pattern is applied to the whole batch, the sub-models generated during training may not be sufficient, which raises the perplexity.

  \section{Conclusion}\label{sect:conclusion}
Accelerating DNN training is difficult because it is nontrivial to leverage the sparsity of DNN in the dense matrix-multiplication.
In this work, we propose a novel approach to eliminate the unnecessary multiplication and data access by replacing the traditional random dropout with an approximate random dropout. The two classes of dropout patterns can avoid the divergence issue in GPU, reduce the scale of the matrix, and thus gain significant improvement on the energy-efficiency with marginal decline of the model performance. The proposed SGD-based search algorithm can guarantee the dropout rate of single neurons or synapse is equivalent to the conventional dropout, as well as the convergence and accuracy of the models. In general, the training time can be reduced by $20\%-77\%$ and $19\%-60\%$ when dropout rate is 0.3-0.7 on MLP and LSTM, respectively. Moreover, higher speedup rate is expected on a larger DNN. The proposed method has been wrapped as an API and integrated into Caffe. The speedup can be much higher if the proposed method can be integrated into the cuBLAS Library.
  \bibliographystyle{ieeetr}
  
  \bibliography{reference}

  \end{document}